%% file: main.tex
\documentclass[wcp,table,xcdraw]{jmlr}
\usepackage{xcolor}
\usepackage{graphicx, soul}
\usepackage{capt-of}
\usepackage{lipsum}
\usepackage{tikz}
\usepackage{enumitem}
\usepackage{comment}
\usepackage{amsmath,amssymb,pifont,fontawesome} %
\usepackage{color}
\usepackage{wrapfig}
\usepackage[accsupp]{axessibility} 
\usepackage{bm}
\usepackage{longtable}% for long tables

\usepackage{booktabs}

\usepackage{boldline} 
\usepackage{mathtools}
\usepackage{capt-of,etoolbox}

\usepackage{dsfont}

\def\ie{\emph{i.e}.}

\def\wrt{w.r.t.}
\def\g{\mathcal{G}_\omega}
\def\f{\mathcal{F}_\theta}
\def\h{\mathcal{H}_\psi}
\def\wt{\mathrm{W}_t}
\def\ws{\mathrm{W}_s}
\def\zs{\mathrm{Z}_s}
\def\zt{\mathrm{Z}_t}
\def\xs{\mathrm{X}_s}
\def\xt{\mathrm{X}_t}
\newcommand{\1}[1]{\mathds{1}\left[#1\right]}
\newcommand{\norm}[1]{\left\lVert#1\right\rVert}
\DeclareMathOperator*{\argmax}{arg\,max}
\DeclareMathOperator*{\argmin}{arg\,min}
\newcommand{\name}[0]{\texttt{CATTAn}}
\newcommand{\sftta}[0]{\texttt{FTTA}}
\newcommand{\tent}[0]{\texttt{TENT}}
\newcommand{\ptent}[0]{\texttt{TENT+}}
\newcommand{\ip}[0]{\texttt{IP}}
\newcommand{\shot}[0]{\texttt{SHOT}}
\makeatletter
\let\Ginclude@graphics\@org@Ginclude@graphics 
\makeatother

\jmlrvolume{}
\jmlrworkshop{Preprint}

\author{\Name{Kowshik Thopalli} \Email{kthopall@asu.edu}\\
   \addr Arizona State University
   \AND
   \Name{Pavan Turaga} \Email{pturaga@asu.edu}\\
   \addr Arizona State University
   \AND
   \Name{Jayaraman J. Thiagarajan} \Email{jjayaram@llnl.gov}\\
   \addr Lawrence Livermore National Laboratory
   }
\title[Domain Alignment Meets Fully Test-Time Adaptation]{Domain Alignment Meets Fully Test-Time Adaptation}

\begin{document}

\maketitle

\begin{abstract}
A foundational requirement of a deployed ML model is to generalize to data drawn from a testing distribution that is different from training. A popular solution to this problem is to adapt a pre-trained model to novel domains using only unlabeled data. In this paper, we focus on a challenging variant of this problem, where access to the original source data is restricted. While fully test-time adaptation (FTTA) and unsupervised domain adaptation (UDA) are closely related, the advances in UDA are not readily applicable to TTA, since most UDA methods require access to the source data. Hence, we propose a new approach, CATTAn, that bridges UDA and FTTA, by relaxing the need to access entire source data, through a novel deep subspace alignment strategy. With a minimal overhead of storing the subspace basis set for the source data, CATTAn enables unsupervised alignment between source and target data during adaptation. Through extensive experimental evaluation on multiple 2D and 3D vision benchmarks (ImageNet-C, Office-31, OfficeHome, DomainNet, PointDA-10) and model architectures, we demonstrate significant gains in FTTA performance. Furthermore, we make a number of crucial findings on the utility of the alignment objective even with inherently robust models, pre-trained ViT representations and under low sample availability in the target domain. 
\let\thefootnote\relax\footnote{This work was performed under the auspices of the U.S. Department of Energy by the Lawrence Livermore National Laboratory under Contract No. DE-AC52-07NA27344, Lawrence Livermore National Security, LLC.}
\end{abstract}

\input{intro}

\input{source_free_tta}
\input{method}

\input{results}

\input{conclusion}

%\acks{Acknowledgements should go at the end, before appendices and references. You can uncomment this for the camera-ready version on paper acceptance.}

%\bibliographystyle{plain}
\small{
\bibliography{acml22}
}

\end{document}

%% file: intro.tex
\section{Introduction}
When the assumption that the training and testing data are drawn from the same distribution is violated, the performance of supervised models can drop drastically~\citep{torralba2011unbiased}. However, in practice, a deployed model is expected to generalize under unknown shifts in the data distribution (e.g., from synthetic to real). Consequently, understanding and improving the generalization of models under such shifts has become an active area of research~\citep{cycada,ganin2016domain,deng2018image}. This problem appears under a variety of formulations in the literature, including domain adaptation~\citep{BenDavid2006AnalysisOR}, domain generalization~\citep{wang2018deep}, few-shot adaptation~\citep{triantafillou2021learning}, and adversarial robustness~\citep{chen2020adversarial}.

In this paper, we focus on unsupervised, fully test-time adaptation (\texttt{TTA}), where a deployed model is adapted using unlabeled data from the target domain, without assuming access to the original source data. This is a practically useful setting, since enabling access to source data during model deployment requires a large memory footprint for common datasets (e.g., ImageNet) and can also lead to shortcomings related to privacy and data usage rights. Existing \texttt{TTA} approaches can be organized based on a) whether data from the \textit{source} domain can be accessed during adaptation; b) which parameters of the source model are updated; and c) whether data from the \textit{target} domain is labeled or unlabeled. A closely related problem is unsupervised domain adaptation (\texttt{UDA}), which attempts to anticipate and adapt for distribution shifts between the labeled source data and unlabeled target data. Despite significant advances in UDA over the last decade, state-of-the-art solutions for \texttt{TTA} do not utilize explicit alignment objectives. This motivates our approach, \name~(\underline{C}alibrate-by-\underline{A}ligning for \underline{T}est \underline{T}ime \underline{A}daptatio\underline{n}) wherein we show that, by leveraging the latent space geometry, we can relax the requirement of source data access, and enable geometric alignment between source and target data at test-time. While our method requires access to source data in the form of basis vectors of a subspace spanned by the source features, it does not store the loadings (or coefficients). Consequently, this neither affects the memory overhead (the basis set requires less than 2 MB of storage in comparison to several GBs of training data) nor compromises the privacy needs, since state-of-the-art deep inversion methods~\citep{iresnet,dreaming,reversible} cannot effectively recover the training data using only features from later layers of a network, let alone with only the subspace basis. Using extensive empirical studies on several standard 2D image and 3D point cloud benchmarks, for the first time, we find that including unsupervised alignment in the cost function leads to significant performance gains over existing fully test-time adaptation methods.
% In both cases, we demonstrate that \name~outperforms state-of-the art \texttt{TTA} models. 
\vspace{0.1in}
\begin{figure}[t]
    \centering
    \includegraphics[width=\textwidth]{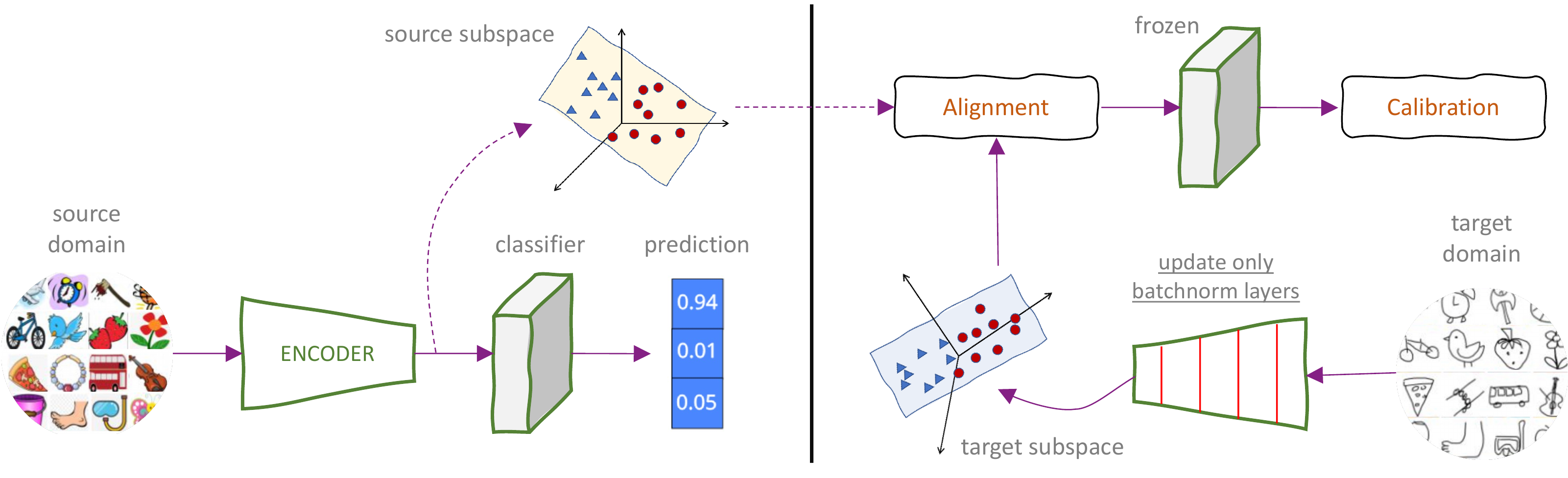}
    \caption{An overview of the proposed approach that incorporate subspace-based feature alignment for fully test-time adaptation. At test time, we only assume access to the trained source model and the subspace approximation of source latent features.}
    
    \label{fig:my_label}
    \vspace{-0.3in}
\end{figure}
    
\noindent \textbf{Contributions:}

\noindent (i) We propose a new test-time adaptation approach \name~to \ul{bridge UDA and \texttt{TTA}, while not requiring access to full source data};

\noindent (ii) We introduce a simple, post-hoc strategy to perform a distribution shift check after the model is already adapted to the target. Through this simple detector, we show that \ul{we can recover the source domain performance even after the model is adapted}; 

\noindent (iii) We perform rigorous empirical studies on \ul{large-scale vision benchmarks (ImageNet, DomainNet, OfficeHome,PointDA-10) and network architectures (ResNet50, ViT)}; 

\noindent (iv) Our codes will be publicly released \url{https://github.com/kowshikthopalli/CATTAn}.

\vspace{0.1in}
\noindent \textbf{Results:}

\noindent (i) \name~produces SoTA results on all benchmarks, outperforming TENT~\citep{Tent}, SHOT~\citep{SHOT}, as well as the recent~\citep{InputTrans} -- ImageNet-C (+$2.1\%$), Office-home (+$2.2\%$) and Office-31 (+$1.7\%$); 

\noindent (ii) To demonstrate the generality of our approach, we also conducted experiments on PointDA-10, a widely adopted 3D point cloud benchmark and observed that \name~improves over existing \texttt{TTA} baselines by $+3.4\%$. 

\noindent (iii) We conduct, for the first time, a \texttt{FTTA} experiment on the large-scale DomainNet~\citep{domainnet} dataset, based on self-supervised representations from the recent ViT-based masked autoencoders~\citep{mae}. We find that \name~produces a boost of $1.1\%$ over the best-performing \texttt{TTA} baseline, and matches the performance of a state-of-the art \texttt{UDA} approach~\citep{roy2021curriculum}; 

\noindent (iv) We find that the proposed geometric alignment objective is beneficial even when the target sample size is limited or when the source model was obtained via robust training~\citep{hendrycks2020augmix}.

%% file: source_free_tta.tex
\section{Fully Test-Time Adaptation}
\label{sec:sfta}
% \begin{enumerate}
%     \item confidence maximization as the core principle for adaptation under distribution shifts
%     \item TENT (entropy based)
%     \item SHOT (pseudo-labeling)
%     \item Input Transformation preprocessing
% \end{enumerate}
Our goal is to improve the generalization of a model trained on the source dataset $\{(\mathrm{x}_s,y_s)\} \in \mathcal{D}_s$ to examples from the target domain $\{(\mathrm{x}_t)\} \in \mathcal{D}_t$ through adaptation under the following conditions -- \underline{c1}: $\mathcal{D}_s \neq \mathcal{D}_t$; \underline{c2}: both source and target domain share the same set of labels; \underline{c3}: examples from $\mathcal{D}_t$ are not labeled; and \underline{c4}: there is no access to original source data samples during adaptation. %, \textit{i.e.}, $y_s, y_t \in \mathcal{C}$

While this work focuses on unsupervised, fully test-time adaptation, a broad class of formulations have been considered in the literature for adapting models under distribution shifts. A popular formulation is conventional transfer learning, which first pre-trains a \textit{source model} using data from $\mathcal{D}_s$, and uses labeled examples from $\mathcal{D}_t$ to perform end-to-end fine-tuning or partial adaptation of selected layers in the source network~\citep{donahue2014decaf,yosinski2014transferable}. In contrast, unsupervised domain adaptation (\texttt{UDA}) jointly infers domain-invariant representations for both labeled source and unlabeled target domain examples, such that they both can utilize a shared classifier. Similarly, \citeauthor{TTT} introduced a test-time training (\texttt{TTT}) protocol based on an auxiliary rotation angle prediction task, which also uses labeled source and unlabeled target examples.

Motivated by the need for source-free adaptation protocols, \citeauthor{SHOT} proposed SHOT that can effectively repurpose a source model, without requiring access to the original source data. Several variants of this approach have been proposed in the literature~\citep{nrc,adaptivesfda,huang2021model} and all of them rely on end-to-end fine-tuning, which can be a bottleneck in fully test-time adaptation (limited data as well as need for fast adaptation). Hence, recent methods such as \texttt{TENT}~\citep{Tent} and \texttt{IP}~\citep{InputTrans} update only the batch normalization layers of the source model.

\begin{table}[h]
\vspace{-0.2in}
\caption{Comparing \name~to existing \sftta~approaches. \textcolor{gray}{Conf. Max.: conditional entropy/NLL, CB: class balance loss, BN: batchnorm.}}
\label{tab:comp}
\renewcommand{\arraystretch}{0.9}
\resizebox{\linewidth}{!}{
\begin{tabular}{c|cccc|ccc}
\multicolumn{1}{c}{\cellcolor[HTML]{9FC5E8} SFTTA Methods} & \multicolumn{4}{c}{\cellcolor[HTML]{FBBC04}Losses}                                                                                          & \multicolumn{3}{c}{\cellcolor[HTML]{34A853}Updates}                     \\
                                                    & Conf. Max. & CB           & pseudo lab. & Geom. Align. & BN params.            & I/P Trans. & Align. Layer \\
\hline
Tent                                                & \faCheck                              & \multicolumn{1}{l}{\ding{53}}     &    \ding{53}                               &         \ding{53}                                & \faCheck                    & \ding{53} &      \ding{53}                                        \\
Tent+                                               & \faCheck                              & \faCheck                    &  \ding{53}                                 &        \ding{53}                                 & \faCheck                    &  \ding{53}   &        \ding{53}                                 \\
SHOT                                                & \faCheck                              & \faCheck                    & \multicolumn{1}{c}{\faCheck}          & \ding{53}                                        & \faCheck                    &        \ding{53}                      &     \ding{53}           \\
IP                                                  & \faCheck                              & \faCheck                    &      \ding{53}                             &          \ding{53}                               & \faCheck                    & \faCheck               &\ding{53}    \\ \hline
Ours                            & \faCheck         & \faCheck&           \ding{53}                        & \faCheck                                   & \faCheck&          \ding{53}     &       \faCheck                \\
\bottomrule
\end{tabular}
}
\vspace{-0.2in}
\end{table}

\paragraph{\textbf{Methodological gaps and Proposed Work.}} We begin by noting that the entropy minimization or other diversity promoting optimization strategies widely adopted by existing \texttt{TTA} methods can be viewed as calibrating predictions from a pre-trained classifier under distribution shifts~\citep{DIRT}. Furthermore, due to the source-free training assumption, they do not leverage any domain alignment objectives. 
Our work is aimed at closing this methodological gap by incorporating explicit domain alignment strategies from \texttt{UDA} into fully test-time adaptation. 
In particular, we employ a novel deep subspace alignment strategy to align the target and source subspaces during adaptation. This modification incurs only a negligible memory burden when compared to \texttt{TENT} (storing the basis vectors of a low-rank subspace). Table \ref{tab:comp} shows how \name~compares to existing \sftta~approaches.
% Through this alignment objective \name~not only achieves better target prediction calibration, but also obtains significantly improved generalization under challenging distribution shifts at test-time. We describe our method in detail in Section~\ref{sec:method}.

%% file: method.tex
\section{Proposed Approach}
\label{sec:method}
% While our approach is aimed at bridging \texttt{UDA} methods and \sftta, it does not require access to complete source data and incurs negligible memory/computational overheads compared to state-of-the-art methods. 
% Specifically, we propose to incorporate an explicit domain alignment objective based on a novel deep subspace alignment (DSA) strategy. Our approach allows end-to-end training and attempts to balance between aligning target and source data distributions, and optimizing the calibration objective.
% Though one might explore other choices for the alignment strategy, e.g., subspace distributional alignment~\citep{sda} or CORAL~\citep{coral}, they typically incur larger memory overheads compared to our approach; for example, storing a covariance matrix of dimensions $D \times D$ as opposed to a low-rank subspace basis of size $D\times d$ ($d<<D$).

As described in Section~\ref{sec:sfta}, our goal is to adapt a source model $\f$ with parameters $\theta$ to a (unlabeled) target domain at test-time. We express $\f \coloneqq \g \circ \h$, as a composition of a feature extractor $\g$ with parameters $\omega$ and a classifier $\h$ with parameters $\psi$ (\ie, $\theta \coloneqq \omega \cup \psi$). During adaptation, the classifier model $\h$ is frozen and only the target features are suitably modified. 
% \noindent\textbf{Notation:}
% We denote through $\zs \in \mathbb{R}^{n_s \times D}$ and $\zt \in \mathbb{R}^{n_t \times D}$ the matrix of latent features where $n_s,n_t$ are the number of samples in source and target domains respectively.
% $\ws,\wt$ denote the basis vectors (principal components) of the source and target subspaces respectively.

\subsection{Geometric Alignment Regularization}
\label{sec:GAR}
Upon training $\f$ on the source dataset, we extract the latent features for source data $\mathrm{Z}_s=\g(\xs)$ where $\zs \in \mathbb{R}^{n_s \times D}$ and $n_s$ is the number of source samples. We then compute a low-dimensional linear subspace with the basis $\mathrm{W}_s \in \mathbb{R}^{D \times d}$ that spans the source features $\mathrm{Z}_s$ using principal component analysis (PCA). Here, $D$ denotes the ambient dimensionality of the latent space and $d$ is the subspace dimension. For test-time adaptation, our approach stores this pre-computed basis set $\mathrm{W_s}$ in addition to the learned model parameters.

In order to introduce an alignment objective between the source and target features, we first extract features for the target data $\xt$ \ie, $\mathrm{Z}_t=\g(\xt)$ and then perform a $d-$dimensional subspace approximation to obtain the corresponding basis $\wt$. Note, $\ws^T \ws = \mathbb{I}$ and $\wt^T\wt = \mathbb{I}$, where $\mathbb{I}$ is the identity matrix. The classical subspace alignment (\texttt{SA}) process estimates the transformation matrix $\Phi$ that aligns $\ws$ and $\wt$:
\begin{equation}
    \label{eq:SAobjective}
    \vspace{-0.1in}
    {\Phi}^* = \argmin_{\Phi} \norm{\wt {\Phi} - \ws}_F^2,
\end{equation}where, $\norm{.}_F$ denotes the Frobenius norm.
The solution to this objective can be obtained in closed form~\citep{Subspace_alignment} as
\vspace{-0.05in}
\begin{equation}
    \label{eq:globalsoln}
        \vspace{-0.1in}
        {\Phi}^* = (\mathrm{W}_t)^{\top} \mathrm{W}_s.
\end{equation}\texttt{SA} then projects $\zs$ onto the source subspace as $\zs \ws$, and the target features $\zt$ onto aligned co-ordinate system (also referred to as the source-aligned target subspace) as $\zt \wt \Phi$. However, na\"ive linear subspace alignment is known to be insufficient for modern datasets with large domain shifts. Hence, \name~uses deep subspace alignment (\texttt{DSA}) that addresses two main challenges: First, we equip \texttt{DSA} with the capability of re-utilizing the source classifier while performing alignment. To this end, we re-project the source-aligned target features into the ambient space as $\bar\wt = \wt\Phi^* = \wt (\wt)^T\ws$ and solve
\begin{alignat}{1}
{\hat{\mathrm{Z}}_t}^* &=  \argmin_{\hat{\mathrm{Z}}_t} \norm{\hat{\mathrm{Z}}_t \mathrm{W}_s -\hat{\mathrm{Z}}_{t} \bar{\mathrm{W}}_t}_{F}^2=\argmin_{\hat{\mathrm{Z}}_t} \norm{\hat{\mathrm{Z}}_t \mathrm{W}_s -\hat{\mathrm{Z}}_{t}  \mathrm{W}_t (\mathrm{W}_t)^{\top} \mathrm{W}_s}_{F}^2,
\end{alignat}where $\hat{\mathrm{Z}}_t^*$ denotes the modified target features. The solution to this optimization is \begin{align}
    \label{eq:reproj}
    {\hat{\mathrm{Z}}_t}^* &= \mathrm{Z}_t\mathrm{W}_t{\Phi}^*\mathrm{W}_s^{\top}.
\end{align}Second, since the eventual goal is not optimal feature alignment but to maximally improve the performance of the model on target data, we include prediction calibration objectives. In such a setting, one can no longer obtain a closed-form solution for ${\Phi}^*$. As a result, \name~uses the following subspace alignment cost in its objective:
\begin{equation}
\label{eq:loss_sa}
    \mathcal{L}_{\Phi}=\norm{\mathrm{W}_t {\Phi} - \mathrm{W}_s}_F^2,
\end{equation}along with objectives that promote well-calibrated predictions on re-projected source-aligned target features $\hat{\mathrm{Z}}_t$ from \eqref{eq:reproj}. To enable end-to-end gradient-based training, we implement \textit{subspace alignment} as a network $\mathcal{A}_{\Phi}(.)$ that parameterizes ${\Phi}$ using a fully connected layer of $d$ neurons without any non-linear activation function or bias \ie~\eqref{eq:reproj} now becomes
 \begin{align}
    \label{eq:new_reproj}
    {\hat{\mathrm{Z}}_t}^* &= \mathcal{A}_{\Phi}(\zt\wt)\ws^{\top}.
\end{align}
Note, we do not use non-linearity because if we include a non-linear activation, $\Phi$ and consequently $\wt \Phi$ will fail to represent linear subspace alignment. Through extensive empirical studies in Section~\ref{sec:results}, we show that, this linear subspace alignment in deep latent spaces is highly effective at improving \sftta~performance.
 \input{algo}

\subsection{Prediction Calibration Objective}
\label{sec:pred_cal}
Calibrating the target predictions using methods such as conditional entropy minimization has been the most common objective in test-time adaptation under distribution shifts, which can be defined as
$H(\hat{y}) = -\sum_c p(\hat{y}_c) \log p(y_c)$, where $\hat{y}= \f(\mathrm{x})$ are the predictions for $\mathrm{x}$ obtained using the model $\f$ and $p(\hat{y}_c)$ denotes the probability for sample $\mathrm{x}$ to be assigned to a specific class $c \in \mathcal{C}$. However, it has been found that entropy minimization can lead to vanishing gradients for high-confidence predictions, thus hindering the training process. Hence, we adopt the non-saturating loss function proposed by~\citeauthor{InputTrans}:
% \begin{equation}
% G(p(\hat{y}), p(y^r)) = - \sum_c p(y^r_c) \log \frac{p(\hat{y}_c)}{\sum_{i \neq c} p(\hat{y}_i)}.
% \end{equation} 
% Formally, the likelihood ratio loss can be computed as follows:
\begin{alignat}{1}
\label{eq:HLR}
\nonumber \mathcal{L}_{lr}(p(\hat{y}))&=-\log \bigg(\frac{p(\hat{y}_{c^*})}{\sum_{i \neq c^*} p(\hat {y}_i)} \bigg)=-\log \bigg(\frac{e^{\hat{y}_{c^*}}}{\sum_{i \neq c^*} e^{\hat{y}_i}}\bigg)=-\hat{y}_{c^*} + \log \sum_{i \neq c^*} e^{\hat{y}_i},
\end{alignat}where $c^* = \argmax p(\hat{y})$. Since this likelihood ratio loss increases the gradient amplitude for high confidence predictions, this is found to be superior to entropy. 

\noindent\textbf{Class Balance Loss:} We also include a popular class diversity objective  $\mathcal{L}_{CB}$ to avoid trivial solutions that are biased towards a subset of the classes, since we perform adaptation using only unlabeled data. $\mathcal{L}_{CB}$ is implemented as the binary cross-entropy between the mean prediction from the network over a mini-batch and an uniform prior distribution.

\noindent\textbf{Overall Objective:} 
The overall objective of \name~is a combination of the alignment cost $\mathcal{L}_{\Phi}$, the prediction calibration term $\mathcal{L}_{lr}$ and the class balance loss $\mathcal{L}_{CB}$:
\begin{equation}
\label{eq:overall_objective}
    \mathcal{L}=  \lambda_{lr}\mathcal{L}_{lr} +  \mathcal{L}_{\Phi} +\lambda_{cb} \mathcal{L}_{CB},
\end{equation}where the penalties $\lambda_{lr}$, $\lambda_{cb}$ are hyper-parameters, the choice of which are not very sensitive as we discuss in our analysis (Sec.~\ref{sec:analysis}).
\vspace{-0.1in}
\subsection{Algorithm}
\label{sec:algo}
\noindent \textbf{Initialization Phase:} Similar to \texttt{TENT}~\citep{Tent}, our method first collects the affine transformation parameters $\left\{\gamma_{l,m}, \beta_{l,m}\right\}$ for each normalization layer $l$ and channel $m$ in the source  model. The remaining parameters $\theta \setminus \left\{\gamma_{l,m}, \beta_{l,m}\right\}$ are not updated during adaptation. As described in Section~\ref{sec:GAR}, our method computes the target features $\zt$ and fits a subspace to obtain $\wt$. We then initialize the deep subspace alignment layer $\mathcal{A}_{\Phi}$ with its weights initialized to $\Phi^*$ from \eqref{eq:globalsoln}.

\noindent\textbf{Adaptation and Termination:} 
In the forward pass, the outputs of the feature extractor $\g$ are transformed through  $\mathcal{A}_{\Phi}$, re-projected using \eqref{eq:new_reproj} and are passed to the classifier. We optimize for the parameters using the objective in \eqref{eq:overall_objective}. We repeat this process for the pre-specified number of epochs. We detail our approach in Algorithm~\ref{algo:algo}. 

\noindent\textbf{Estimating subspace dimension:}
To select the optimal subspace dimension $d$, a hyper-parameter in our approach, we adopt the theoretical stability\begin{wrapfigure}{r}{0.4\textwidth}
\centering
\vspace{-0.1in}
    \includegraphics[width=0.38\textwidth]{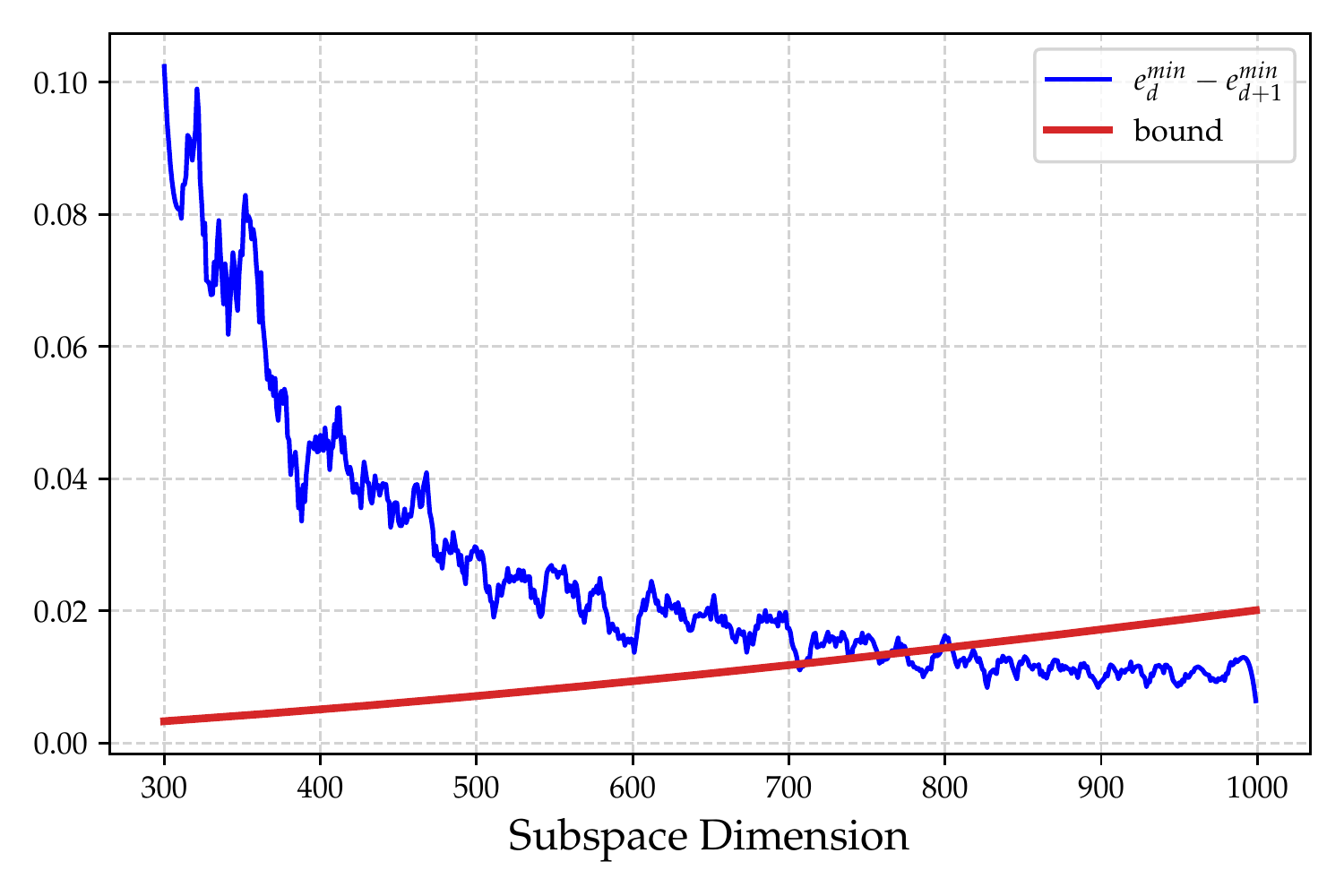}
\vspace{-0.2in}
  \caption{\small{Estimating subspace dimensionality using \eqref{eq:bound} for the A$\rightarrow$C setting from OfficeHome. The lower bound is plotted in red and the difference between consecutive eigenvalues in blue.}}
\vspace{-0.4in}
  \label{fig:subspace_dim}
\end{wrapfigure}
result from~\citep{Subspace_alignment} and modify it for the \sftta~setting. For a given $\delta > 0$ and $\epsilon >0$, we select the maximum subspace dimension $d$ such that
\begin{equation}
\label{eq:bound}
\left(e_{d}^{\min }-e_{d+1}^{\min }\right) \geq\left(1+\sqrt{\frac{\ln 2 / \delta}{2}}\right)\left(\frac{16 d^{3 / 2} }{\epsilon \sqrt{n_{t}}}\right),
\end{equation} where $e_{d}$ represents the $d^{th}$ eigenvalue and $n_t$ denoting the number of samples in target domain.
This theoretical bound gives us a selection rule for picking an optimal $d$.
Given the principal components for both source and target datasets, and the corresponding eigenvalues, we compute the  deviations $e_{d} -e_{d+1}$, $\forall d$, for both source and target data. Through \eqref{eq:bound}, we then obtain a stable solution $d << D$ for a given $\delta$ and $\epsilon$. In our experiments, we set $\delta=0.1$ and $\epsilon=10^6$. For example, we plot the values of the bound and 
$\left(e_{d}^{\min }-e_{d}^{\min }\right)$ w.r.t. to subspace dimension for the $A \rightarrow C$ case from OfficeHome in Figure~\ref{fig:subspace_dim} and we pick the value of $d=800$.

%% file: algo.tex
% \RestyleAlgo{boxruled}
\begin{algorithm2e}[t]
\SetAlgoLined
\DontPrintSemicolon
\caption{Proposed algorithm for fully test-time adaptation}
	\label{algo:algo}
	\textbf{Input}:
	Source-model $\f$; Source subspace $\ws$; target data $\xt$\;
	\textbf{Initialize}: $\lambda_{lr},\lambda_{cb}, n_{iter}$; \ul{Freeze classifier $\h$}; Collect affine transformation parameters $\left\{\gamma_{l,m}, \beta_{l,m}\right\}$ for each normalization layer $l$ and channel $m$ in $\g$\\
	\textbf{Adaptation}:\;
	$\zt=\g \xt$; \tcp{\textcolor{gray}{compute target features}}
	$\wt \leftarrow \text{PCA}(\zt)$ \tcp{\textcolor{gray}{compute target subspace}}
	Compute $\Phi^*$ using~\eqref{eq:globalsoln}\;
	Initialize the weights of $\mathcal{A}_{\Phi}$ with $\Phi^*$
	
 	\For{iter \textbf{in} $n_{iter}$}{
			$\zt= \g(\xt)$; \tcp{\textcolor{gray}{compute features for target samples}}
			${\hat{\mathrm{Z}}_t}=\mathcal{A}_{\Phi}(\zt\wt)\ws^{\top}$ following \eqref{eq:new_reproj} \tcp{\textcolor{gray}{project, align and re-project}}
			$\bm\hat{\mathrm{y}}_t = \f(\bm\hat{\mathrm{Z}}_t)$ \tcp{\textcolor{gray}{compute predictions for aligned target data}}
			$\mathcal{L}= \lambda_{lr} \mathcal{L}_{lr} +  \mathcal{L}_{\Phi} +\lambda_{cb} \mathcal{L}_{CB}$ using \eqref{eq:overall_objective} \tcp{\textcolor{gray}{compute overall objective}}
            Update alignment $\mathcal{A}_{\Phi}$ and parameters $\left\{\gamma_{l,m}, \beta_{l,m}\right\}$ of $\g$ $\wrt ~\mathcal{L}$ 
	}
\textbf{Output:} $\mathcal{A}^{*}_{\Phi},\g^{*},\f^{*}$

\end{algorithm2e}

%% file: results.tex
\section{Experiments}
\label{sec:results}
\noindent \textbf{List of Experiments:}
In Table~\ref{tab:loe}~we provide the details of different experiments we conducted, their goals and the models and datasets for a quick reference. In addition, we provide a discussion on hyper-parameter choices and ablations of our method. 

\begin{table}[h]

\renewcommand{\arraystretch}{1.2}
\resizebox{\textwidth}{!}{
\begin{tabular}{|c|c|c|c|}
\toprule
Evaluation & Model & Datasets & Section \\
\midrule
Utility of~\name~for large-scale corruptions  & ResNet-50 & ImageNet $\rightarrow$ ImageNet-C &  sec~\ref{sec:ImageNet} \\
Performance of \name~for common UDA benchmarks & ResNet-50 & OfficeHome, Office-31 &  sec~\ref{sec:uda_benchmarks} \\
\name~for 3D point-cloud classification & PointNET & PointDA-10 & sec~\ref{sec:point_cloud}\\
Efficacy of \name~with pre-trained ViT embeddings & MAE with ViT as backbone & DomainNet &  sec~\ref{sec:mae} \\
Impact of target sample sizes & Resnet-50 & OfficeHome &  sec~\ref{sec:lowdata} \\
 Extending \name~to recover source performance & ResNet-50 & OfficeHome &  sec~\ref{sec:source_perf} \\
Impact of robust training on \name~& Robust ResNet-50 & OfficeHome &  sec~\ref{sec:robust} \\
\bottomrule
\end{tabular}}
\caption{List of experiments.}
\label{tab:loe}
\end{table}

\noindent\textbf{Datasets:} We evaluate \name~using standard \texttt{UDA} datasets along with a robustness benchmark, ImageNet $\rightarrow$ ImageNet-C. 
(i) The \ul{OfficeHome}~\citep{officehome} dataset is comprised of 15,500 images from $65$ classes, where the images belong to $4$ different domains; (ii) The \ul{Office-31} dataset~\citep{saenko2010adapting} contains 4110 images from $31$ classes and represents three different domains; (iii) \ul{DomainNet}~\citep{domainnet} is a large scale \texttt{UDA} benchmark with $500$K images from $6$ domains with $345$ classes each; (iv) ImageNet $\rightarrow$ \ul{ImageNet-C}~\citep{hendrycks2018benchmarking} is a challenging corruption robustness benchmark that includes $15$ types of synthetic corruptions with $5$ severity levels; and (v) \ul{PointDA-10} is the first $3$D point-cloud benchmark specifically designed for domain adaptation and comprises point-clouds belonging to $10$ categories across $3$ domains. In total, it contains approximately $27.7$K training and $5.1$K test samples. \\
\noindent\textbf{Models:} As our method operates under the \sftta~setting, any arbitrary pre-trained model can be used. 
We experiment with the publicly available (pre-trained) \ul{Resnet-50}~\citep{he2016deep} model for evaluation on the ImageNet-C benchmark, and the modified Resnet-50 architecture from~\citep{SHOT} for the \texttt{UDA} benchmarks. Furthermore, we also experiment with a \ul{vision transformer(ViT)}~\citep{vit}-based encoder (trained using masked auto-encoders~\citep{mae}) finetuned on the ImageNet dataset.
For PointDA-10, we use the \ul{PointNET}~\citep{pointnet} backbone proposed in PointDAN~\citep{pointdan}. As this model has only a single $1$D BN layer, we extend the architecture with $4$ additional $2$D BN layers (\textit{i.e.}, after the convolutional layers).\\   
\noindent\textbf{Baselines:} We consider the following state-of-the-art \sftta~methods for evaluation: (i) \tent~\citep{Tent}; (ii) \ptent, a variant of \tent that includes the class-balance loss defined in Section~\ref{sec:pred_cal}; (iii) The recent \ip~\citep{InputTrans} approach that includes a learnable input transformation module (convolutional layers) to correct for the shifts; and (iv) \texttt{SHOT}~\citep{SHOT} that uses a pseudo-labeling based optimization strategy for test-time adaptation. Note that, the model architectures and the training protocols (e.g., update only BN layers) were fixed to be the same for all methods.

\noindent\textbf{Metrics:}
We use the accuracy and empirical calibration error (ECE)~\citep{guo2017calibration} metrics for our evaluation. \\
\noindent\textbf{Setup:}
For all \texttt{UDA} benchmarks, following standard practice, we considered each of the domains as source and adapted the source model to each of the target domains at test-time independently. We implemented \name~in PyTorch and used the Adam optimizer with learning rate $1e-4$ and set the batch size to $64$. All experiments were repeated thrice with three different random seeds, and we report the average performance. Moreover, in cases where validation sets were not specified, we performed a $90-10$ random split, and used the validation split to select hyper-parameters. For \ip~and \name, we set $\lambda_{lr} = 0.025$ in all our experiments. We implemented \tent~and \ip~and generated results for Office-31, OfficeHome and DomainNet datasets, as their performance on these datasets have not been reported in their respective papers. We adapt \tent~from the publicly available codebase~\footnote{https://github.com/DequanWang/\tent~}, while we re-implemented \ip, since their code was not publicly released.
Following the strategy outlined in sec~\ref{sec:algo}, we picked the subspace dimensionality $d$ for our experiments.
While \tent~has been found to be useful for online adaptation (single epoch), \citeauthor{Tent} found that performing the adaptation for more epochs consitently leads to better performance. Hence, in our experiments, we performed $5$ epochs of adaptation for all methods.

\input{tables/imagenet}

\section{\sftta~Performance on $2$D and $3$D Benchmarks}
\subsection{ImageNet-C Benchmark}
\label{sec:ImageNet}
In Table~\ref{tab:imagenet}, we report the performance of our proposed method, along with the baselines, on ImageNet-C at the highest severity level 5. It can be observed that the proposed method improves over \tent, \ptent~and \shot~by 6\% points and \ip~by 2\% points respectively. Among the baselines, \ip~performs the best - this can be attributed to the additional trainable input transformation module, which is typically well-suited for handling pixel-level corruptions. However, by not adopting an explicit alignment objective and using only the prediction calibration process to guide the adaptation, \ip~produces lower performance than \name, which does not employ any image-space transformation.

\subsection{\texttt{UDA} Benchmarks}
\label{sec:uda_benchmarks}
\input{tables/officehome}

\input{tables/office31}

We demonstrate the efficacy of our method under large distribution shifts found in typical UDA problems by performing experiments with OfficeHome and Office-31 datasets. The comparative results for these two datasets can be found in Tables \ref{tab:office-home} and \ref{tab:office31} respectively.  
Similar to the observations from the previous experiment, \name~consistently performs better than the existing \sftta~baselines. On OfficeHome, \name~improves upon \tent, \ip~and \shot~by $3.8,2.2$ and $2.8\%$ points respectively, while on Office-31 \name~produces gains of $2.4, 1.8$ and $3.04\%$ points. Interestingly, while the input transformation module proposed in \ip~is useful with pixel-level corruptions, it is not able to achieve invariance to the large semantic shifts that occur in typical domain adaptation benchmarks. As a result, the performance of \ip~tends to be similar to that of \ptent. In contrast, the latent subspace alignment strategy adopted by \name~produces large performance gains over \ptent. As we consider more complex datasets with large diversities between domains going forward, we compare our method against the more general and stronger baseline \ptent. 
\subsection{$3$D point-cloud Dataset}
\label{sec:point_cloud}
As discussed earlier, our latent space alignment strategy is applicable to different model architectures or data modalities. In order to demonstrate this, we experimented with a recent $3$D point-cloud classification DA benchmark (PointDA-$10$). As shown in Table~\ref{tab:pointcloud}, the adaptation performance of \name~is significantly superior to \tent~and \ptent~by $3.4\%$ and $2.8\%$ points respectively (averaged across 6 experiments). Especially, in cases such as Model$\rightarrow$Shape and Scan$\rightarrow$Model, \name~improves upon \ptent~by more than $6\%$ points while matching its performance in challenging settings such as Model$\rightarrow$Scan. This clearly evidences the effectiveness of our approach across different problem settings. 
\input{tables/pointda}

\section{\name~with Pre-Trained ViT Embeddings}
\label{sec:mae}
\begin{figure}
    \centering
    \includegraphics[width=\textwidth]{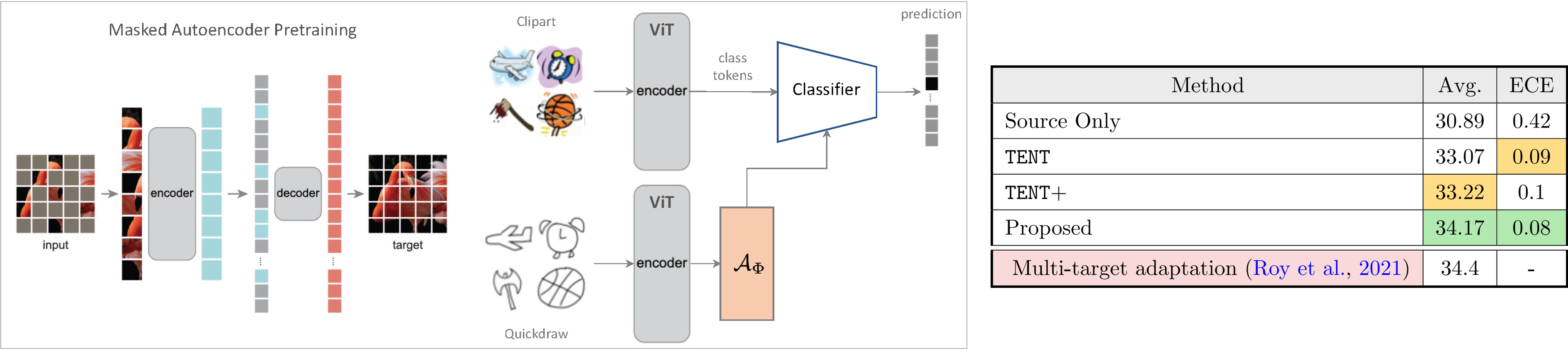}
    \caption{Implementing \name~with pre-trained representations from Masked Autoencoders. The table shows the \sftta~performance on \textbf{DomainNet} with representations from MAE~\citep{mae}.}
    \vspace{-0.3in}
    \label{fig:mae}
\end{figure}

As Transformer-based solutions such as vision transformers (ViT)~\citep{vit} and masked auto encoders (MAE)~\citep{mae} are becoming increasingly popular and achieve state-of-art performance in solving vision problems, it is imperative to understand the efficacy of our alignment strategy on feature representations obtained from such large-scale pre-trained transformer encoders. To this end, we consider the encoder from MAE~\citep{mae} fine-tuned on ImageNet as our feature extractor~\footnote{https://github.com/facebookresearch/mae/blob/main/FINETUNE.md}. As illustrated in Figure~\ref{fig:mae}, MAE first masks a large portion of the image and attempts to reconstruct the complete image from the masked image. Once trained via this self-supervision, the encoder is then fine-tuned with ImageNet data. We then freeze the encoder, obtain source features (class tokens) $\zs$ and perform PCA to obtain the basis $\ws$. Note that, we do not fine-tune the ViT with source domain data, but instead use it as an off-the-shelf feature extractor. A source model, which is comprised of a single MLP layer with batch normalization and a linear classifier layer, is then constructed and trained using $\zs$. During adaptation with unlabeled target data, we extract features $\zt$ from the frozen encoder and obtain subspace basis vectors $\wt$. Using the optimization procedure of \name, the batch normalization parameters and the subspace alignment module $\mathcal{A}_\Phi$ are then estimated.

In this experiment, we used the large-scale DomainNet~\citep{domainnet} benchmark. Given that there are $6$ domains in this dataset, we conducted a total of 30 test-time adaptation experiments, where we consider one domain as source and the other domains as target. In Figure \ref{fig:mae}, we report the average performance obtained on this benchmark. It can be seen that \name~improves upon \ptent~by almost $1\%$ (averaged over 30 experiments), thus indicating that even under this challenging setting, the alignment objective plays an important role. For comparison, we also include the result from a state-of-the-art multi-target domain adaptation approach~\citep{roy2021curriculum}, which trains a Resnet-101 model end-to-end with combined source and target data. Our results show that, with a powerful feature encoder, simple test-time adaptation with \name~can produce similar performance. This clearly emphasizes the improved representational power of modern pre-training strategies, as well as the efficacy of \name~in aligning disparate domains even without accessing the entire source data.

% updating affine transformation parameters along with an alignment cost, \ptent and \name~matches the performance of a model that has been end-to-end trained with access to full source and target data. 
\section{Analysis}
\label{sec:analysis}
\subsection{Behavior of \name}
\paragraph{Role of different loss terms:} In Table~\ref{tab:comp}, we highlight the differences between
the different SoTA \sftta~baselines and our method. 
Our extensive empirical study with multiple SoTA benchmarks and these baselines clearly show that the proposed geometric alignment is critical for the reported performance gains.

\paragraph{Choice of $\lambda_{cb}$:} 
Using OfficeHome~\citep{officehome} dataset, we study the sensitivity of \name~\wrt~change in the value of $\lambda_{cb}$. As can be evidenced from Figure~\ref{fig:analysis_cattan}(a), for values greater than $0.4$ the performance of \name~is stable with respect to changes in $\lambda_{cb}$. In our experiments, we fixed $\lambda_{cb}$ to be $1.0$.

\subsection{Impact of Target Sample Sizes}
\label{sec:lowdata}
Next, we study the impact of \name~under varying sample complexity in the target
\begin{figure}
  \begin{center}
    \includegraphics[trim =0 0 0 0 0,clip,width=0.85\textwidth]{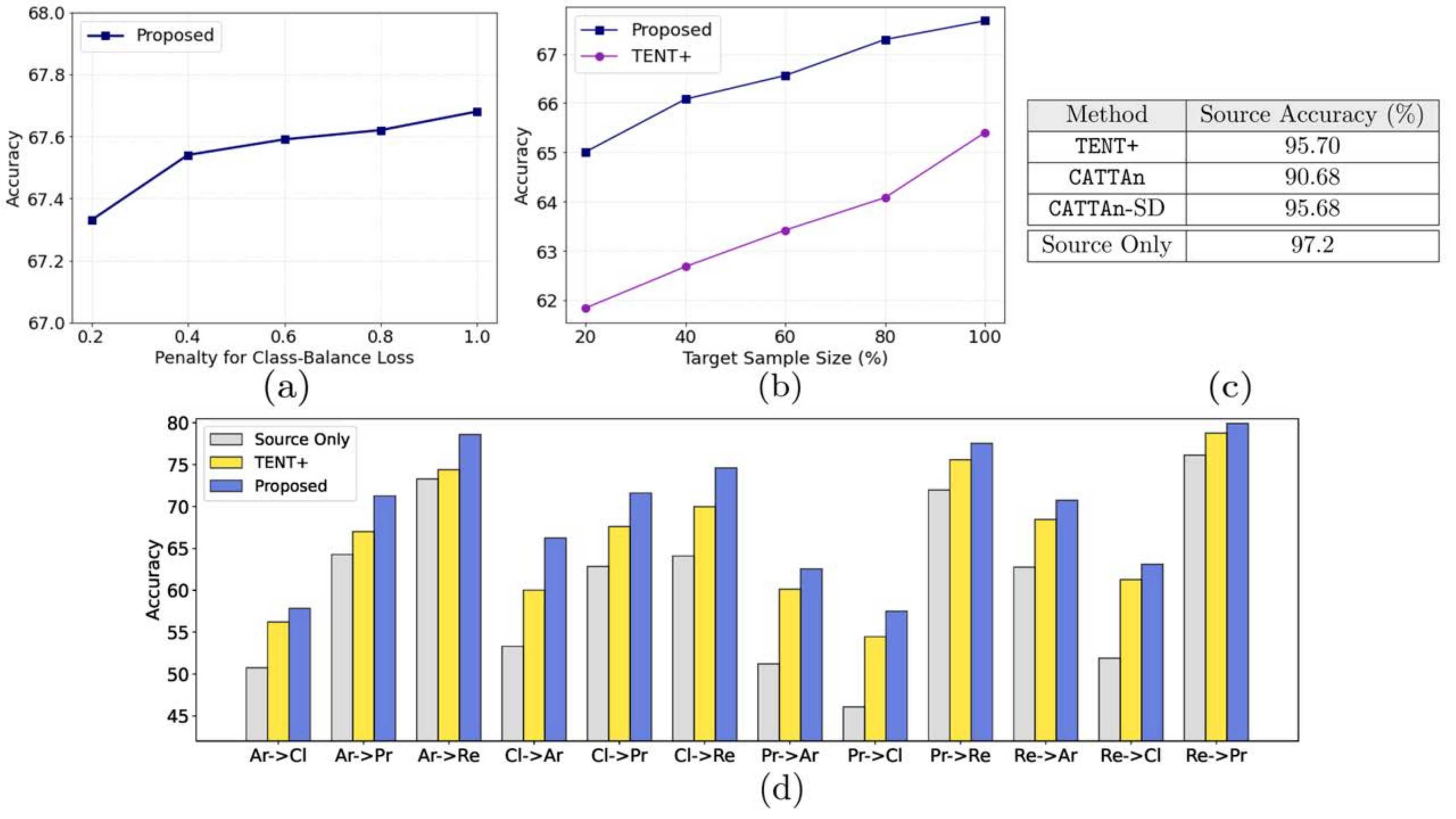}
  \end{center}
    \vspace{-0.3in}
  \caption{Analysis of \name~- (a) Performance of \ptent~and \name~for varying target sample sizes;
  (b) Performance of \name~on the OfficeHome dataset with varying values of class-balance penalty $\lambda_{cb}$;
  (c) Enabling \name~to recover performance on the original source data through a simple test-time shift detection mechanism; (d) \sftta~performance with a robust source model -- \name~improves upon the baselines, this indicating that the proposed approach provides non-trivial invariances not captured by a robust model.}
\vspace{-0.2in}
  \label{fig:analysis_cattan}
\end{figure} domain. While \name~is not expected to be very effective for online adaptation with a small batch (due to the poor quality of subspace fit in high-dimensions with sparse examples), we make an interesting finding that, even at significantly reduced sample sizes, \name~produces superior adaptation performance than strong baselines such as \ptent. In Figure~\ref{fig:analysis_cattan}(b), we plot the performance of \ptent~and \name~at different sample sizes. 
Unsurprisingly, the performance drops as the target size decreases but importantly, the drop in performance of \name~is less severe than that of \ptent. 
\subsection{Extending \name~to Recover Source Performance}
\label{sec:source_perf}
\input{expt_source_rec}

\subsection{Impact of Robust Training on \name}
\label{sec:robust}
Having empirically established the effectiveness of \name~in achieving state-of-the-art results using standard models such as ResNet-50, a natural question to then ask is if the alignment cost still helps when we consider an inherently robust model instead. To answer this, we conducted experiments with a robust ResNet50 model~\citep{hendrycks2021many}, trained with DeepAugment and Augmix~\citep{hendrycks2020augmix} strategies. 
From Figure~\ref{fig:analysis_cattan}(d), we first notice that the source-only performance is improved by more than $1.3\%$ points when compared to the standard model, which indicates that a robust model already captures at least some of the invariant properties. Furthermore, we notice that both \ptent~and \name~improve upon no adaptation performance significantly, with average accuracies $66.1\%$ and $69.29\%$ respectively. With a performance boost of more than $3\%$ points than \ptent, \name~clearly evidences the importance of including an alignment objective for \sftta~even with a robust model. We also remark that, since most standard robustness training paradigms do not include large distribution shifts such as the diverse shifts encountered in UDA datasets, alignment techniques such as \name~provide complementary benefits.

%% file: tables/imagenet.tex
\begin{table}[t]
\caption{Results on all 15 corruptions of \textbf{ImageNet-C} benchmark at the highest severity level-5 using standard Resnet50. Through the inclusion of an alignment objective, \name~improves significantly upon existing baselines.}
\renewcommand*{\arraystretch}{1.3}
\resizebox{\textwidth}{!}{
\begin{tabular}{V{3}l|c|c|c|c|c|c|c|c|c|c|c|c|c|c|c|cV{3}}
\hlineB{3}
\rowcolor[HTML]{EBEBEB} 
\multicolumn{1}{V{3}c|}{\cellcolor[HTML]{EBEBEB}{Method}} &\multicolumn{1}{c|}{\cellcolor[HTML]{EBEBEB}gauss}&\multicolumn{1}{c|}{\cellcolor[HTML]{EBEBEB}SHOT}&\multicolumn{1}{c|}{\cellcolor[HTML]{EBEBEB}impulse}&\multicolumn{1}{c|}{\cellcolor[HTML]{EBEBEB}defocus}&\multicolumn{1}{c|}{\cellcolor[HTML]{EBEBEB}glass}&\multicolumn{1}{c|}{\cellcolor[HTML]{EBEBEB}motion}&\multicolumn{1}{c|}{\cellcolor[HTML]{EBEBEB}zoom}&\multicolumn{1}{c|}{\cellcolor[HTML]{EBEBEB}snow}&\multicolumn{1}{c|}{\cellcolor[HTML]{EBEBEB}frost}&\multicolumn{1}{c|}{\cellcolor[HTML]{EBEBEB}fog}&\multicolumn{1}{c|}{\cellcolor[HTML]{EBEBEB}bright}&\multicolumn{1}{c|}{\cellcolor[HTML]{EBEBEB}contrast}&\multicolumn{1}{c|}{\cellcolor[HTML]{EBEBEB}elastic}&\multicolumn{1}{c|}{\cellcolor[HTML]{EBEBEB}pixel}&\multicolumn{1}{c|}{\cellcolor[HTML]{EBEBEB}jpeg}& \multicolumn{1}{cV{3}}{\cellcolor[HTML]{EBEBEB}Avg.} \\ \hline
\rowcolor[HTML]{FFFFFF} 
Source Only                                                                             &4.7 & 5.4 & 4.7 & 15.1 & 8.9 & 13.1 & 22.8 & 15.6 & 20.3 & 22.7 & 55.6 & 4.4 & 14.8 & 23.1 & 33.3 & 17.6                            \\ \hline
\rowcolor[HTML]{FFFFFF} 
\tent                                                                                   &16.54 & 18.6 & 16.64 & 16.78 & 17 & 28.72 & 42.66 & 39.72 & 34.8 & 51.78 & 66.16 & 14.32 & 47.4 & 50.84 & 40.56 & 33.50                              \\ \hline
\rowcolor[HTML]{FFFFFF} 
\tent+                                                                                    &16.96 & 19.1 & 17.3 & 17.1 & 17.56 & 29.22 & 42.82 & 40.04 & 35.4 & 51.82 & 65.82 & 15.78 & 47.64 & 50.88 & 40.68 & 33.87                            \\ \hline
\rowcolor[HTML]{FFFFFF} 
\texttt{SHOT}                                                                                      &17.34 & 21.12 & 20 & 18.42 & 20.06 & 33.41 & 43.04 & 38.65 & 36.99 & 54.33 & \cellcolor[HTML]{b1e9b0}67.54 & 16.78 & 51.59 & 51.75 & 43.35 & 35.62                           \\ \hline
\rowcolor[HTML]{FFFFFF} 
\texttt{IP}                                                            &\cellcolor[HTML]{FFDD86}23.94 & \cellcolor[HTML]{FFDD86}26.88 & \cellcolor[HTML]{FFDD86}25.06 & \cellcolor[HTML]{FFDD86}23.2 & \cellcolor[HTML]{FFDD86}22.62 & \cellcolor[HTML]{FFDD86}36.28 & \cellcolor[HTML]{FFDD86}48.7 & \cellcolor[HTML]{FFDD86}46.58 & \cellcolor[HTML]{FFDD86}39.44 & \cellcolor[HTML]{FFDD86}56.08 & \cellcolor[HTML]{b1e9b0}67.58 & \cellcolor[HTML]{FFDD86}18.6 & \cellcolor[HTML]{FFDD86}53.1 & \cellcolor[HTML]{FFDD86}55.58 & \cellcolor[HTML]{FFDD86}48.76& \cellcolor[HTML]{FFDD86}39.49                          \\ \hline
\rowcolor[HTML]{FFFFFF} 
Proposed                                                     & \cellcolor[HTML]{b1e9b0}26.02 & \cellcolor[HTML]{b1e9b0}30.4 & \cellcolor[HTML]{b1e9b0}28.82 & \cellcolor[HTML]{b1e9b0}26.06 & \cellcolor[HTML]{b1e9b0}26.7 & \cellcolor[HTML]{b1e9b0}41.02 & \cellcolor[HTML]{b1e9b0}49.34 & \cellcolor[HTML]{b1e9b0}47.46 & \cellcolor[HTML]{b1e9b0}39.42 & \cellcolor[HTML]{b1e9b0}57.0 & \cellcolor[HTML]{FFDD86}66.52 & \cellcolor[HTML]{b1e9b0}23.88 & \cellcolor[HTML]{b1e9b0}54.4 & \cellcolor[HTML]{b1e9b0}57 & \cellcolor[HTML]{b1e9b0}50.48 & \cellcolor[HTML]{b1e9b0}41.63  \\ \hlineB{3}
\end{tabular}
}
\label{tab:imagenet}
\vspace{-0.2in}
\end{table}

%% file: tables/officehome.tex
\begin{table}[t]
\caption{Results on the \textbf{OfficeHome Dataset} obtained using Resnet50. Our approach improves upon existing \sftta~baselines. Interestingly, \ip~and \ptent~baselines perform similarly, indicating that the input transformation module in \ip~is not effective at undoing larger domain shifts.}
\renewcommand*{\arraystretch}{1.1}
\resizebox{\textwidth}{!}{
\begin{tabular}{V{3}l|c|c|c|c|c|c|c|c|c|c|c|c|c|cV{3}}
\hlineB{3}
\rowcolor[HTML]{EBEBEB} 
\multicolumn{1}{V{3}c|}{\cellcolor[HTML]{EBEBEB}{Method}} & {A  $\rightarrow$ C} & {A  $\rightarrow$ P} & {A  $\rightarrow$ R} & {C  $\rightarrow$ A} & {C  $\rightarrow$ P} & {C  $\rightarrow$ R} & {P  $\rightarrow$ A} & {P  $\rightarrow$ C} & {P  $\rightarrow$ R} & {R  $\rightarrow$ A} & {R  $\rightarrow$ C} & {R  $\rightarrow$ P}& {Avg.}& {ECE} \\ \hline
\rowcolor[HTML]{FFFFFF} 
Source Only                                                   &       43.73                  &     65.35                        &      72.94                       &      52.62                      &     61.07                         &    64.77                         &     51.17                        &      40.53                       &   73.01                          &   64.65                          &    45.25                         &77.27  & 59.36&       0.56                     \\ \hline
\rowcolor[HTML]{FFFFFF} 
\tent                                                          & 47.88 & 65.98 & 73.26 & 58.76 & 65.94 & 68.07 & 60.16 & 47.31 & 75.4 & \cellcolor[HTML]{FFDD86}70.83 & 53.95 & 78.73 & 63.85  &  0.09                            \\ \hline
\rowcolor[HTML]{FFFFFF} 
\tent+                                                          &51.48 & \cellcolor[HTML]{FFDD86}69.07 & 74.39 & 59.21 & 67.52 & 69.43 & \cellcolor[HTML]{FFDD86}60.49 & 50.1 & \cellcolor[HTML]{FFDD86}76.34 & \cellcolor[HTML]{FFDD86}70.83 & 56.29 & 79.82 & \cellcolor[HTML]{FFDD86}65.41      & \cellcolor[HTML]{FFDD86}0.07                            \\ \hline
\rowcolor[HTML]{FFFFFF} 
\texttt{SHOT}                                                          &      50.61 & 68.69 &\cellcolor[HTML]{FFDD86}74.71 & 58.34 & 67.63 & \cellcolor[HTML]{FFDD86}70.07 & 57.73 & 49.14 & \cellcolor[HTML]{FFDD86}76.38 & 69.47 & 54.89 & \cellcolor[HTML]{FFDD86}79.88 & 64.795 &  \cellcolor[HTML]{b1e9b0}0.06                           \\ \hline
\rowcolor[HTML]{FFFFFF} 
\texttt{IP}                                                            &  \cellcolor[HTML]{FFDD86}52.16 & \cellcolor[HTML]{FFDD86}69.09 & 74.57 & \cellcolor[HTML]{FFDD86}59.7 & \cellcolor[HTML]{FFDD86}67.79 & 69.31 & 60.2 & \cellcolor[HTML]{FFDD86}50.63 & 75.72 & 70.58 & \cellcolor[HTML]{FFDD86}56.38 & 79.61 & \cellcolor[HTML]{FFDD86}65.47  &    \cellcolor[HTML]{FFDD86}0.07                       \\ \hline
\rowcolor[HTML]{FFFFFF} 
Proposed                                                      &  \cellcolor[HTML]{b1e9b0}52.81 & \cellcolor[HTML]{b1e9b0}73.89 & \cellcolor[HTML]{b1e9b0}77.07 & \cellcolor[HTML]{b1e9b0}61.93 & \cellcolor[HTML]{b1e9b0}71.12 & \cellcolor[HTML]{b1e9b0}72.94 & \cellcolor[HTML]{b1e9b0}61.89 & \cellcolor[HTML]{b1e9b0}52.35 & \cellcolor[HTML]{b1e9b0}79.05 & \cellcolor[HTML]{b1e9b0}72.11 & \cellcolor[HTML]{b1e9b0}56.68 & \cellcolor[HTML]{b1e9b0}80.27 & \cellcolor[HTML]{b1e9b0}67.68 &   0.08                         \\ \hlineB{3}
\end{tabular}
}
\vspace{-0.2in}
\label{tab:office-home}
\end{table}

%% file: tables/office31.tex
\begin{table}[t]
\centering
\caption{Adaptation results for \textbf{Office31 Dataset} obtained using Resnet50. We observe that \name~consistently improves upon state-of-the-art \sftta~approaches.}
\renewcommand*{\arraystretch}{1.1}
\resizebox{0.63\textwidth}{!}{
\begin{tabular}{V{3}l|c|c|c|c|c|c|c|cV{3}}
\hlineB{3}
\rowcolor[HTML]{EBEBEB} 
\multicolumn{1}{V{3}c|}{\cellcolor[HTML]{EBEBEB}{Method}} & {A  $\rightarrow$ C} & {A  $\rightarrow$ P} & {A  $\rightarrow$ R} & {C  $\rightarrow$ A} & {C  $\rightarrow$ P} & {C  $\rightarrow$ R}& {Avg.}& {ECE} \\ \hline
\rowcolor[HTML]{FFFFFF} 
Source Only~&   81.12                          & 74.47                             &     61.34                        &    94.34                         &62.62                             &   97.39   &78.54 &   0.6                         \\ \hline
\rowcolor[HTML]{FFFFFF} 
\tent                                                         & 82.13 & 85.16 & 68.83 & 97.48 & 62.94 & \cellcolor[HTML]{FFDD86}99.8 & 82.72 &    0.11                 \\ \hline

\rowcolor[HTML]{FFFFFF} 
\tent+                                                         &82.33 & \cellcolor[HTML]{FFDD86}85.66 & \cellcolor[HTML]{FFDD86}69.72 & 97.61 & 65.03 & \cellcolor[HTML]{FFDD86}99.8 & \cellcolor[HTML]{FFDD86}83.35 & \cellcolor[HTML]{FFDD86}0.10                     \\ \hline

\rowcolor[HTML]{FFFFFF} 
\texttt{SHOT}                                                          &    80.72 & 82.64 & 67.59 & 97.23 & 64.54 & \cellcolor[HTML]{FFDD86}99.8 & 82.08 & 0.18                  \\ \hline
\rowcolor[HTML]{FFFFFF} 
\texttt{IP}                                                            &\cellcolor[HTML]{FFDD86}82.73 & 85.28 & 69.12 & \cellcolor[HTML]{FFDD86}97.99 & \cellcolor[HTML]{FFDD86}65.35 & \cellcolor[HTML]{b1e9b0}100 & \cellcolor[HTML]{FFDD86}83.41 &\cellcolor[HTML]{b1e9b0}0.07                   \\ \hline
\rowcolor[HTML]{FFFFFF} 
Proposed                                                      &   \cellcolor[HTML]{b1e9b0}85.54 & \cellcolor[HTML]{b1e9b0}86.29 & \cellcolor[HTML]{b1e9b0}72.88 & \cellcolor[HTML]{b1e9b0}98.62 & \cellcolor[HTML]{b1e9b0}67.59 & \cellcolor[HTML]{FFDD86}99.8 & \cellcolor[HTML]{b1e9b0}85.12 & \cellcolor[HTML]{b1e9b0}0.07                     \\ 
 \hlineB{3}
\end{tabular}
}
\label{tab:office31}
\end{table}

%% file: tables/pointda.tex
\begin{table}[h]
\centering
\label{tab:pointcloud}
\caption{Adaptation results on the \textbf{PointDA-10}, a $3$D point cloud classification benchmark. We observe that the proposed approach provides an improvement of over $1.5\%$, thus evidencing its generality across model architectures and data modalities.}
\renewcommand*{\arraystretch}{1.3}
\resizebox{0.9\textwidth}{!}{
\begin{tabular}{|
>{\columncolor[HTML]{FFFFFF}}c |
>{\columncolor[HTML]{FFFFFF}}c |
>{\columncolor[HTML]{FFFFFF}}c |
>{\columncolor[HTML]{FFFFFF}}c |
>{\columncolor[HTML]{FFFFFF}}c |
>{\columncolor[HTML]{FFFFFF}}c |
>{\columncolor[HTML]{FFFFFF}}c |c|}
\hline

\cellcolor[HTML]{EBEBEB} & \cellcolor[HTML]{EBEBEB}Model$\rightarrow$Shape & \cellcolor[HTML]{EBEBEB}Model$\rightarrow$Scan & \cellcolor[HTML]{EBEBEB}Shape$\rightarrow$Model & \cellcolor[HTML]{EBEBEB}Shape$\rightarrow$Scan & \cellcolor[HTML]{EBEBEB}Scan$\rightarrow$Model & \cellcolor[HTML]{EBEBEB}Scan$\rightarrow$Shape & \cellcolor[HTML]{EBEBEB}Mean  \\ \hline
% Source Only~\citep{pointdan}             & \cellcolor[HTML]{FFFFFF}43.1        & 17.3                               & 40                                  & 15                                 & 33.9                               & 47.1                               & 32.73                         \\ \hline
Source Only  & \cellcolor[HTML]{FFFFFF}52.1        & 15.74                              & 51.32                               & 12.79                              & 38.82                              & 52.14                              & 37.15                         \\ \hline
\tent                   & 54.69                               & \cellcolor[HTML]{B1E9B0}23.38      & 51.82                               & \cellcolor[HTML]{B1E9B0}28.4       & 38.1                               & \cellcolor[HTML]{FFDD86}53.44      & 41.97                         \\ \hline
\ptent                    & \cellcolor[HTML]{FFDD86}56.2        & \cellcolor[HTML]{B1E9B0}23.32      & \cellcolor[HTML]{FFDD86}52.33       & \cellcolor[HTML]{FFDD86}27.41      & \cellcolor[HTML]{FFDD86}42.48      & 53.71                              & \cellcolor[HTML]{FFDD86}42.58 \\ \hline
\name                 & \cellcolor[HTML]{B1E9B0}62.41       & \cellcolor[HTML]{FFDD86}22.83                              & \cellcolor[HTML]{B1E9B0}54.44       & 27.11                              & \cellcolor[HTML]{B1E9B0}48.58      & \cellcolor[HTML]{B1E9B0}57.07      & \cellcolor[HTML]{B1E9B0}45.41 \\ \hline
\end{tabular}}
\end{table}

%% file: expt_source_rec.tex
\input{algo_source_recovery}

From our experiments with several DA benchmarks, we find that using an explicit alignment objective leads to significantly improved test performance. However, this comes at a cost of a drop in accuracy for samples from the original source domain after adaptation. This is an inherent challenge with methods that include an explicit alignment during adaptation. As a toy example, consider the case where the samples from target domain are rotated versions of samples from source domain by a certain angle. In this case, ideally $\mathcal{A}_{\Phi}$ would be the matrix that will undo this rotation. However, at test-time, if the same $\mathcal{A}_{\Phi}$ is applied to source data, the classifier will fail, as this creates a new domain shift of rotating by twice the angle. To address this issue, we propose a post-hoc strategy to determine if a test sample belongs to the target domain or OOD (\textit{i.e.}, from the source domain). If the sample is OOD, $\mathcal{A}_{\Phi}$ is no longer applicable and we replace it with $\mathcal{A}_{\Phi}= \mathbb{I}$ (identity matrix). Note that, this approach for OOD detection does not require access to source data and hence is applicable with any off-the-shelf model.

Our post-hoc mechanism is inspired by the recent results in measuring generalization gap using inconsistencies between multiple hypotheses in a deep ensemble~\citep{jiang2021assessing}. Let us assume that we have $K$ different hypotheses $\{\f^{1} \cdots \f^{K}\}$ for the prediction function that we want to approximate. We rely on the inter-hypothesis consistency to check for distribution-shifts. The intuition here is that, an OOD sample has a higher risk of having inconsistent predictions across the different hypotheses. Specifically, for each sample we average the prediction probabilities from $K-1$ models and assign the label corresponding to the class that has the highest probability. We compare this prediction against the prediction of $K^{th}$ model. We repeat this for all $k$ in $K$ \ie~
\begin{equation}
\label{eq:qvalue}
  q(x) =\sum^{K}_{k=1} \1{ \frac{1}{K-1} \bigg(argmax\bigg(\sum^{K}_{i,i\neq k}\f^{i}(x)\bigg)\bigg)== argmax(\f^{k}(x))}  
\end{equation} We finally obtain the normalized score for the inconsistency $\bar{q}(x)$ as $\bar{q}(x)= \frac{q(x)}{K}$. Intuitively, larger the $q$ value for a given sample at test-time, higher is the likelihood for it to be drawn from the target distribution. To facilitate this comparison, we compare $\bar{q}(x)$ against a user defined threshold $\tau$ \ie, if $\bar{q}(x) < \tau$ then use $\mathcal{A}_{\Phi}=\mathbb{I}$. We set the $\tau$ to be 0.75 in our experiments. 

We adopt the following approach to construct the multiple source hypotheses. The first hypothesis construction follows the procedure explained in Section~\ref{sec:GAR}. For the second hypothesis, we fit another target subspace but only to two-thirds of the target samples that had the lowest confidence values and recompute the target subspace using them. Note, even with this new target subspace, the adaptation process uses the entire target data. In other words, the diversity in the hypotheses arises mainly from the different target subspace fits. In our experiments, we set $K=3$. 
As can be seen from Figure~\ref{fig:analysis_cattan}(c) this simple test can effectively recover the accuracy on the source dataset. We also note that, this test can be applied to any \sftta~method (e.g., to avoid applying the input transformation to a source sample in \ip).

%% file: algo_source_recovery.tex
% % \RestyleAlgo{boxruled}

% \begin{algorithm2e}[t]
% \SetAlgoLined
% \DontPrintSemicolon
% \caption{Modified t}
% 	\label{algo:algo_recovery}
% 	\textbf{Input}:
% 	Source-model $\f$,

% 	\textbf{Initialize}: $\Phi$ using \eqref{eq:globalsoln}; number of members in ensemble $K$; confidence threshold $\beta$; q score threshold $\tau$;\\
%  	\textbf{Building an ensemble:}\;
%  	\For{k \textbf{in} $K$}{
% 			updated $\Phi^{*}, f^{k*} \leftarrow $ Algorithm~\ref{algo:algo}($\{\mathrm{Z}_t\},\wt$)\tcp{Adapt the model}
%             compute confidence scores $\{\max p(\hat{y})\}$ for target data\;
%             sorted = sort($\{\max p(\hat{y})\})$ \tcp{samples that have less confidence}
%             Outliers= find(sorted $< \beta$ ) \;%\tcp{take lowest 66\%}
%           % $\{\mathrm{Z}_t\} \leftarrow \{\mathrm{Z}_t\}$[outliers];\tcp{collect outliers}
%             $\wt \leftarrow \text{PCA}(\text{Outliers})$ \tcp{fit subspace for outliers only}
% 	}   
%     \textbf{OOD detection:}\;
%     \For{ $\mathrm{x}$ \textbf{in} $\{\mathrm{X}_t\}$ }{
%      Compute $q(x)$ using Eq.~\ref{eq:qvalue};
%     %  $q(x) =\sum^{K}_{k=1} \1{ \frac{1}{K-1} \bigg(argmax\bigg(\sum^{K}_{i,i\neq k}f^{i}(x)\bigg)\bigg)== argmax(f^{k}(x))}$\tcp{Eq.~\ref{eq:qvalue}}
%     $\bar{q}= \frac{q}{K}$ \tcp{normalized q score}
%     \eIf{$\bar{q}\ge \tau$}{
%     ID $\leftarrow \mathrm{x}$ \tcp{sample is ID \ie from target domain}
%     }
%     {
%     OOD $\leftarrow \mathrm{x}$ \tcp{sample is OOD \ie from source domain}
%     $\mathcal{A}_{\Phi}=\mathbb{I}$ \tcp{since source we dont apply alignment}
%     }}
% \end{algorithm2e}

%% file: conclusion.tex
\section{Conclusions}
In this work, we explored the benefit of including an alignment objective in test time adaptation. First, we show that we can bridge UDA and TTA solutions without requiring access to complete source data.  Through \name~we incorporated deep subspace alignment and demonstrated the effectiveness of alignment across different benchmarks. Using rigorous empirical studies, we showed that our method consistently improves and outperforms the state-of-the-art methods on several $2$D image and $3$D point-cloud benchmarks. We also showed the effectiveness of the proposed method when we consider a powerful feature extractor such as Vision transformers and robust models. Interestingly, our method is robust at even low sample sizes. We also proposed a novel post hoc algorithm that can be applied after the model is adapted to target domain such that the model is still effective on the source data.  Future extensions of the work include extending to other vision tasks such as semantic segmentation and exploring other alignment techniques which do not require source data.